\documentclass[sigconf]{acmart}

\AtBeginDocument{%
  }

\setcopyright{acmlicensed}

\copyrightyear{2025}
\acmYear{2025}
\setcopyright{acmlicensed}\acmConference[MM '25]{Proceedings of the 33rd ACM International Conference on Multimedia}{October 27--31, 2025}{Dublin, Ireland}
\acmBooktitle{Proceedings of the 33rd ACM International Conference on Multimedia (MM '25), October 27--31, 2025, Dublin, Ireland}
\acmDOI{10.1145/3746027.3755115}
\acmISBN{979-8-4007-2035-2/2025/10}

\begin{document}

\title{Towards Consumer-Grade Cybersickness Prediction: Multi-Model Alignment for Real-Time Vision-Only Inference}

\author{Yitong Zhu}
\email{yzhu162@connect.hkust-gz.edu.cn}
\affiliation{%
  \institution{The Hong Kong University of Science and Technology(Guangzhou)}
  \city{Guangzhou}
  \state{Guangdong}
  \country{China}
}

\author{Zhuowen Liang}
\email{simonliang484@gmail.com}
\affiliation{%
  \institution{The Hong Kong University of Science and Technology(Guangzhou)}
  \city{Guangzhou}
  \state{Guangdong}
  \country{China}
}

\author{Yiming Wu}
\email{wuyi0031@e.ntu.edu.sg}
\affiliation{%
  \institution{Nanyang Technological University}
  \country{Singapore}
}

\author{Tangyao Li}
\email{tli724@connect.hkust-gz.edu.cn}
\affiliation{%
  \institution{The Hong Kong University of Science and Technology(Guangzhou)}
  \city{Guangzhou}
  \state{Guangdong}
  \country{China}
}

\author{Yuyang Wang}
\authornote{Corresponding author}
\email{yuyangwang@hkust-gz.edu.cn}
\affiliation{%
  \institution{The Hong Kong University of Science and Technology(Guangzhou)}
  \city{Guangzhou}
  \state{Guangdong}
  \country{China}
}

\begin{abstract}
Cybersickness remains a major obstacle to the widespread adoption of immersive virtual reality (VR), particularly in consumer-grade environments. While prior methods rely on invasive signals such as electroencephalography (EEG) for high predictive accuracy, these approaches require specialized hardware and are impractical for real-world applications. In this work, we propose a scalable, deployable framework for personalized cybersickness prediction leveraging only non-invasive signals readily available from commercial VR headsets, including head motion, eye tracking, and physiological responses. Our model employs a modality-specific graph neural network enhanced with a Difference Attention Module to extract temporal-spatial embeddings capturing dynamic changes across modalities. A cross-modal alignment module jointly trains the video encoder to learn personalized traits by aligning video features with sensor-derived representations. Consequently, the model accurately predicts individual cybersickness using only video input during inference. Experimental results show our model achieves 88.4\% accuracy, closely matching EEG-based approaches (89.16\%), while reducing deployment complexity. With an average inference latency of 90ms, our framework supports real-time applications, ideal for integration into consumer-grade VR platforms without compromising personalization or performance. The code will be relesed at https://github.com/U235-Aurora/PTGNN.
\end{abstract}



\maketitle

\section{Introduction}\label{sec:intro}

Virtual Reality (VR) has swiftly evolved from a niche technology into a mainstream platform, enabling immersive experiences in gaming, healthcare~\cite{wu2024aienhancedvirtualrealitymedicine}, industrial safety~\cite{VR-based-industries}, and education~\cite{jin2023development, Jiang2024ChemistryVR}. Advances in rendering fidelity, spatial tracking, and real-time interaction~\cite{JEON2023104929} have significantly enhanced the realism and interactivity of VR environments. Despite these advancements, cybersickness, a visually induced variant of motion sickness manifesting as dizziness, nausea, and discomfort, continues to pose a significant barrier to the sustained and widespread adoption of virtual reality (VR) technologies.

As VR expands into consumer markets, there is an increasing demand for lightweight, real-time, and user-adaptive solutions that can effectively predict cybersickness on affordable mobile hardware. However, most existing approaches were developed from limited environments or research settings, thus failing to meet the constraints and usability of consumer-grade VR applications. Prior research has explored diverse strategies, including predictive content adaptation, reduction of visual-vestibular conflicts, and physiological signal modeling. Early efforts primarily focused on visual factors, such as dynamic field-of-view adjustments~\cite{Padmanaban2019FOVReduction} and motion compensation algorithms~\cite{Smith2021VisualVestibularConflict}. More recent studies have turned to physiological signals, with EEG-based models demonstrating promise in capturing neural correlates of discomfort~\cite{Kim2019CybersicknessEEG}.

To better understand why existing methods fall short in real-world applications, we highlight three critical limitations that must be addressed.~(1) \textbf{Device Dependency}:~EEG-based approaches~\cite{Chen2020EEGCybersickness, Jeong2019DeepLearningAlgorithms} rely on high-accuracy devices, increasing cost and limiting portability.~(2) \textbf{Lack of Personalization}:~many content-driven models~\cite{Kim2020VideoDrivenModels} lack personalization, failing to account for individual behavioral traits such as gaze stability or head movement, and thus struggle with cross-user generalization, and~(3) \textbf{Real-Time Constraints}:~deep learning models that process video or high-dimensional signals often incur high latency (typically exceeding 4000ms)~\cite{kundu2023lite}, making them unsuitable for real-time, interactive VR experiences.

These limitations highlight the urgent need for cybersickness prediction frameworks that are not only accurate and personalized but also compatible with commercially available hardware. While recent work has made progress toward this goal, critical gaps remain in achieving a balance between performance and practicality. For instance, Kim et al. \cite{Kim2019CybersicknessEEG} proposed personalized cybersickness modeling by analyzing VR video sequences and EEG signals, achieving 89.16\% accuracy through user-specific EEG spectral patterns. However, dependency on signals like EEG, requiring invasive and expensive hardware and impeding natural VR interactions, renders this approach impractical for consumer-grade deployment. Alternative methods prioritize accessible sensor modalities. Chang et al. \cite{Chang2021PredictCybersicknessGaze} developed a regression model using eight variables to predict cybersickness, explaining 34.8\% of the variance in the total SSQ score. Nevertheless, regression-based models underperform compared to deep learning frameworks, which leverage temporal dependencies in behavioral data — a capability absent in shallow models. Recently, Ramaseri-Chandra and Reza ~\cite{ramaseri2025dynamic} proposed an adaptive system that can leverage real-time data and provide immediate feedback about the user's sickness level. While promising, continuous data streaming increases hardware costs, posing scalability issues for consumer applications.
\begin{figure}[htbp]
  \centering
  \includegraphics[width=1.0\linewidth]{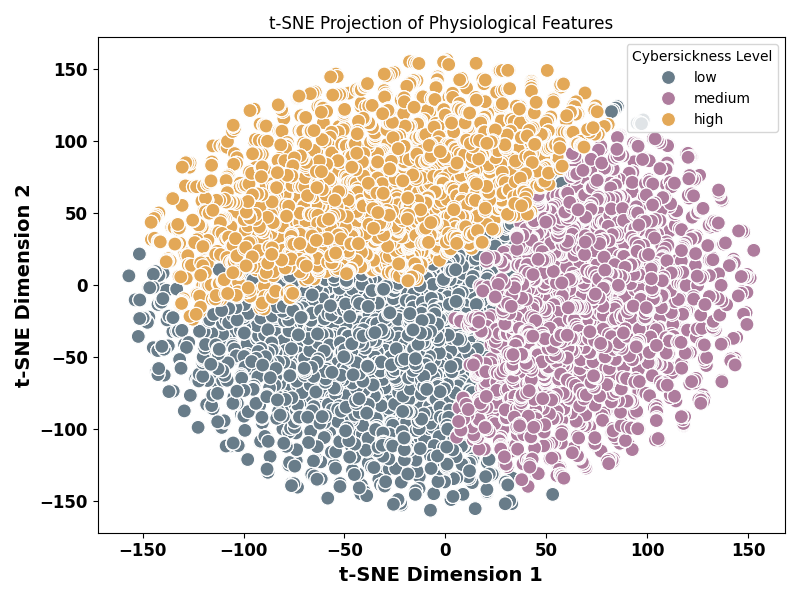} 
  \caption{t-SNE of user responses to a shared VR video}
  \label{fig:replace}
\end{figure}

We aim to design a framework capable of predicting cybersickness based solely on VR video input, thereby improving both predictive performance and consumer accessibility. Individual differences in video perception have been widely observed across EEG and eye-tracking studies~\cite{Wu2024HypergraphMultiModalLLM}, motivating our use of personalized non-invasive embeddings. To achieve this while preserving individual variability and eliminating reliance on specialized hardware such as EEG, we incorporate consumer-grade, non-invasive multimodal signals-including head and eye movements, as well as physiological data such as  EDA (electrodermal activity), BVP (blood volume pulse), and SKT (skin temperature)-to model user-specific traits.  We begin by confirming that, when exposed to the same VR content, participants exhibit significant inter-individual differences in these signals. As illustrated in Fig.~\ref{fig:replace}, users with varying cybersickness levels display clearly separable distributions in the physiological and behavioral feature space. The t-SNE projection reveals distinct clusters corresponding to low, medium, and high cybersickness groups, indicating that these signals encode discriminative characteristics associated with user discomfort. These findings demonstrate that individual cybersickness responses can be effectively captured without EEG, supporting the feasibility of using accessible multimodal data for personalized modeling.

From this observation, we can find that the non-invasive data can capture personalized embeddings instead of invasive signals. To tackle the above challenges, we propose a cross-modal joint training network, \textit{\textbf{MM-PTGNN}}. First, we extract modality-specific embeddings from non-invasive signals. These embeddings are used to construct an adaptive multi-modal graph that models the spatial-temporal relationships between modalities. The Difference Attention module is designed to fuse these diverse features by focusing on their dynamic differences, enhancing information integration from different modalities. The video encoder is trained to learn personalized features through cross-modal alignment with sensor data. This approach enables real-time, personalized prediction solely with VR video, without wearable sensors, making it suitable for consumer-grade applications.

The key contributions of this work are as follows:
\begin{itemize}
\item We design a joint learning framework that aligns video and non-invasive sensor embeddings during training. This enables the video encoder to inherit personalized traits, allowing video-only inference with subject-aware cybersickness prediction.

\item We propose a Difference Attention Module that computes cross-modal attention based on signal differences, enabling effective fusion of multiple non-invasive signals while mitigating baseline noise and enhancing dynamic sensitivity.

\item We have conducted experiments on a multi-modal dataset using only non-invasive signals with video, achieving performance comparable to EEG-based state-of-the-art methods while supporting real-time inference, highlighting its great practicality and scalability for consumer-grade VR applications.
    
\end{itemize}

\section{Related Work}\label{sec:related work}

\subsection{Personal and Content-Driven Cybersickness Analysis}
The perceptual mechanisms of cybersickness are rooted in classical theories such as sensory conflict (incoherence between visual, vestibular, and proprioceptive inputs) and postural instability (failure to maintain dynamic equilibrium)~\cite{riccio1991ecological}. Neurophysiological evidence now bridges these theories to measurable biological responses. Gavgani et al.~\cite{gavgani2018cybersickness} revealed that VR-induced cybersickness triggers elevated hemodynamic activity in the prefrontal cortex (via fNIRS) and suppresses autonomic regulation (e.g., reduced heart rate variability), directly linking sensory mismatch to quantifiable neural and physiological signatures. These findings validate multimodal sensing (e.g., head motion + HRV) as a viable pathway for objective cybersickness quantification.  

However, susceptibility to such sensory conflicts varies substantially across individuals, governed by demographic and experiential factors. Females exhibit 23\% higher symptom severity than males under identical VR exposures~\cite{stanney2020sex}, likely due to hormonal modulation of vestibular sensitivity. Aging populations face increased risk due to natural vestibular degradation~\cite{rebenitsch2016individual}, while frequent VR users develop adaptive tolerance through neural plasticity~\cite{laghari2021veteran}. Such heterogeneity necessitates personalized models that account for biological predispositions and behavioral adaptations. The above personalities will show differences in the head, eye motion, and physiological signatures. 

Beyond individual traits, VR content design critically modulates cybersickness intensity. Fernandes and Feiner~\cite{fernandes2016combating} demonstrated that dynamically restricting the field-of-view (FOV) during high-speed scenes reduces simulator sickness scores, effectively mitigating visual-vestibular conflicts. Conversely, aggressive camera motions, particularly abrupt directional shifts, and mechanical jitter amplify sensory mismatches. Kim et al.~\cite{Kim2020EstimateVRSickness} established a logarithmic correlation ($R^2=0.78$) between camera shake magnitude and subjective discomfort, underscoring the need for motion-stabilized rendering. 

The interdependence of user and content factors remains underexplored. For instance, vestibular-sensitive users may require stricter FOV constraints during camera motions that tolerant users perceive as benign. Our work addresses this gap by jointly modeling user-specific physiological baselines (e.g., head motion stability) and scene kinematic features (e.g., optical flow divergence), enabling adaptive VR systems that dynamically balance immersion and comfort.

\subsection{Deep Learning for Cybersickness Prediction}

Early research on cybersickness prediction predominantly focused on single-modality approaches, each targeting distinct aspects of the problem yet constrained by inherent trade-offs. Video-driven methods, such as Padmanaban et al.~\cite{Padmanaban2018StereoscopicVideos}, leveraged stereoscopic cues (frame disparities, optical flow) to estimate discomfort with an RMSE of 12.00, but their reliance on static scene features overlooked temporal symptom dynamics. Expanding this paradigm, Balasubramanian et al.~\cite{Balasubramanian2019Egomotion} curated a monoscopic video dataset with annotated camera trajectories (shake, velocity, depth), linking aggressive motions to discomfort; however, their user-agnostic design, ignoring individual vestibular sensitivity or prior VR experience, suffered a 19\% accuracy drop on heterogeneous populations. 
This fragmentation reveals a trilemma: high accuracy (EEG), low device dependency (IMU), and content-awareness (video) remain mutually exclusive in single-modality frameworks. Video models encode scene dynamics yet fail to personalize predictions; EEG captures biological fidelity but tethers users to invasive hardware; behavioral tracking enables real-time deployment but lacks physiological granularity. 

In response to the accuracy-accessibility-content trilemma inherent in single-modality frameworks, recent advances pivot toward multimodal integration, strategically combining complementary data streams to mitigate individual weaknesses. Lee et al.~\cite{Lee2019MotionSickness} exemplify this shift by augmenting traditional motion velocity and depth features with eye movement-derived saliency maps, thereby capturing both scene dynamics (optical flow) and user attention patterns (fixation hotspots) — a synergy that reduced RMSE by 22\% compared to video-only models. Similarly, Kim et al.~\cite{Kim2019BrainSignalAnalysis} bridged the hardware dependency gap through a two-stage RNN: EEG signals first encode cognitive load (e.g., frontal alpha asymmetry), which are then fused with real-time scene kinematics (e.g., camera acceleration) in a shared latent space, achieving 89\% accuracy without requiring continuous EEG monitoring.

The quest for consumer-grade practicality further drives sensor-efficient fusion. Islam et al.~\cite{Islam2021IntegratedHMDsSensors} demonstrated that combining native HMD sensors (head motion IMU, eye tracking) and stereoscopic video features (disparity maps) in a unified transformer architecture achieves 91\% accuracy, rivaling EEG-based models while eliminating external biosensors. Beyond minimal sensor setups, Jeong and Han~\cite{jeong2022leveraging} introduced an attention-based fusion framework that harmonizes eye tracking, head movement, and physiological signals (EDA, blood volume pulse, skin temperature), revealing that electrodermal activity (EDA) peaks correlate strongly with vestibular conflicts during sudden FOV shifts. Their model’s dynamic attention weights adaptively prioritize modalities based on symptom severity, reducing false positives by 31\%.

Architectural innovations now push multimodal prediction toward earlier and finer-grained detection. Choi et al.~\cite{choi2024early} leveraged pre-trained large language models (LLMs) to encode temporal dependencies across heterogeneous sensor streams (IMU, heart rate, gaze), achieving early cybersickness prediction with an RMSE of 1.696, 19\% lower than conventional LSTM baselines. However, their LLM-based approach incurs significant computational overhead (about 300ms latency), highlighting a critical trade-off between model complexity and real-time viability. Despite these advances, multimodal frameworks still grapple with the consumer-grade application and the balance of the user-content co-analysis. Our work addresses these gaps by using easier-to-achieve data.

\section{Methodology}
\begin{figure*}[!htbp]
    \centering
    \includegraphics[width=0.99\linewidth]{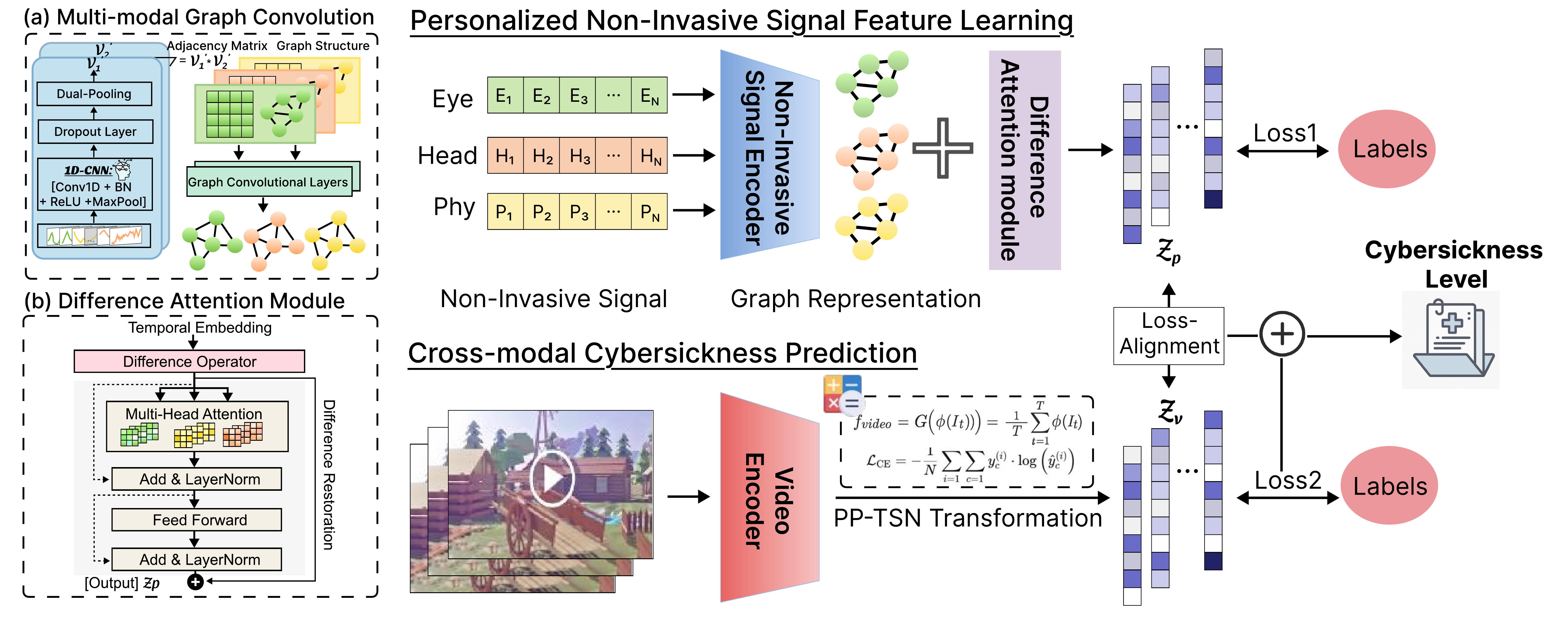}
    \caption{The overall framework of the joint learning work.~(a) Multi-modal Graph Convolution constructs spatial-temporal graphs for each non-invasive modality and learns modality-specific embeddings;~(b) Difference Attention Module incorporates local differential information to enhance attention to dynamic signal variations.}
    \label{fig:overall-framework}
\end{figure*}

To address the aforementioned challenges of balancing real-time deployment and personalized prediction in consumer-grade VR applications, we propose the \textbf{MM}-\textbf{PTGNN}: a Multi-Modal Progressive Temporal Graph Neural Network. The overall architecture is illustrated in Fig.~\ref{fig:overall-framework}. It comprises four main components: a multi-modal graph convolution module, a difference attention module, a video progressive temporal segment network, and a cross-modal alignment module. Given an individual's time-series data, we construct modality-specific graphs for dynamic spatiotemporal representation learning from the multi-modal graph convolution module. A difference attention module further enhances sensitivity to temporal changes while mitigating baseline drift. Video semantics are captured from the video progressive temporal network, and a cross-modal alignment module transfers personalized traits from sensor-derived to video-derived embeddings via supervised learning. In the following, we elaborate on the core components of MM-PTGNN from Section 3.1 to Section 3.4, and describe the loss we use in our network in Section 3.5.

\subsection{Multi-modal Graph Convolution with Multi-modal Embedding Learning}
Multi-modal time series data in our approach has three parts: the eye-tracking motion data, the head-tracking motion data, and the physiological sensor data. Different modalities of the data consist of multiple features that are often closely correlated. For example, in eye-tracking motion, each eye motion may exhibit different features, indicating recognizable spatial patterns of visual attention; in head-tracking motion, different motion parameters (e.g., pitch, yaw, roll angles for head orientation, and linear/angular acceleration for movement dynamics) can capture distinct behavioral signatures tied to attentional shifts, cognitive engagement, or physical fatigue. In physiological data, different sensors can provide complementary information about the state of the cognitive neural system. This motivates us to construct explicit structures to model modal-level and temporal dynamics for individual time series. To this end, we propose a novel multi-modal adaptive graph structure learning approach to model different modalities of dynamics. This approach consists of two steps:~1)~learning multi-modal embeddings,~2)~constructing multi-modal graph structure based on the learned embeddings.

\subsubsection{Multi-modal Embedding Learning} 
To catch the multi-scale spatial and temporal dependency, we propose an embedding learning framework based on the dimension decoupling. All three input modals are fed into the encoder. To learn a compact and informative representation for each modality, we employ a hierarchical 1D convolutional neural network (1D-CNN) to extract temporal patterns from the processed sequences. The network consists of three convolutional layers, each followed by Batch Normalization, ReLU activation, and Max-Pooling to downsample and extract hierarchical temporal features progressively. The output feature embeddings are further processed in later stages for multimodal integration. For instance, given the input tensor of physiological sensor data, $X \in \mathbb{R}^{B \times T \times N \times D}$, where T is the size of the sliding window. The 1D-CNN serves as a modality-specific feature extractor, encoding time-dependent characteristics while reducing noise and redundancy. By leveraging a stack of convolutional layers and pooling operations, the model effectively captures short-term dependencies and compresses information into a lower-dimensional embedding space, followed by:
\begin{align}
\text{Stage 1:} \quad & Y^{(1)} = \text{Dropout}\left(\mathcal{P}\left(\sigma(\text{BN}(W^{(1)} \ast X^{(0)}))\right)\right) \\
\text{Stage 2:} \quad & Y^{(2)} = \text{MaxPool} \left(\sigma(\text{BN}(W^{(2)} \ast Y^{(1)}))\right) \\
\text{Stage 3:} \quad & Y^{(3)} = \text{MaxPool} \left(\sigma(\text{BN}(W^{(3)} \ast Y^{(2)}))\right)
\end{align}
, where $\sigma(\cdot)$ is the ReLU function, $\mathcal{P}$ represents MaxPooing, and $W^{(l)}\in\mathbb{R}^{K\times C_{in}^{(l)}\times C_{out}^{(l)}}$ are the convolutional kernels. The final feature embeddings $E_{eye}$, $E_{head}, E_{phy}\in$$ \mathbb{R}^{B \times T \times N \times D}$ are used for subsequent GNN-based graph modeling.

\subsubsection{Multi-modal Graph Construction} 
To model intra-modality spatial relationships, we construct a fully connected graph over the nodes, parameterized by a learnable adjacency matrix $A \in \mathbb{R}^{N \times N}$. We define a two-layer Graph Convolutional Network (GCN) for each modality. For each modality, we define an undirected graph $G$. The graph convolution operation at time step $t$ is defined as:
\begin{equation}
    E_{t}^{'}=\sigma(AE_{t}W_{1}), Z_{t}=A{H'_{t}W_{2}}
\end{equation}
, where $E_{t} \in \mathbb{R}^{B \times N \times D}$ is the node embedding at time t, and $A$ are learnable weight matrices, and $\sigma(\cdot)$ denotes a ReLU function. 

To promote adaptability and enable data-driven topological learning, the adjacency matrix is initialized as a symmetric matrix and optimized during training. This allows the model to infer latent connectivity patterns across sensor nodes, without relying on predefined topology. After graph convolution, we apply temporal average pooling to aggregate the node-level features across time:
\begin{equation}
Z=\frac{1}{T}\sum_{t=1}^{T}Z_{t}
\end{equation}
Each modality yields an aggregated embedding $Z_{eye}$, $Z_{head}$ , $Z_{phy}$ = $GNN(H^{(i)}, A_{i})$ respectively, and will be input into the difference attention mechanism. 

\subsection{Difference Attention Module}
\label{DAE}
Considering the dynamic sensitivity in the multi-modal signal, to capture more efficient information hidden behind the immediate change of the signal, we propose the difference attention method motivated by the finite difference method~\cite{fan2025medgnnmultiresolutionspatiotemporalgraph}. Different from the traditional attention mechanism, our method computes the weight of the attention by the change rate of the signal(difference features) and enhances the sensitivity in the dynamic mode. At the same time, it helps reduce interference from baseline shift, as supported by findings in~\cite{qiu2023DifferAttention}.

We first employ a feature projection step to align feature dimensions across multiple graphs to integrate heterogeneous graph representations from different modalities (e.g., head motion, eye movement, and physiological data). Given a set of graphs \(\mathcal{G} = \{ G_1, G_2, G_3 \}\), where each graph \(G_i\) contains node features in a different dimensional space \( \mathbb{R}^{C_i} \), we aim to project them into a shared latent space of dimension \( \mathbb{R}^{d} \). The transformation is formulated as follows:  
\begin{equation}
    X_i' = W_i X_i + b_i, \quad \forall i \in \{1,2,3\}
\end{equation}
, where \( X_i \in R^{B \times T \times N_i \times C_i} \) is the original feature matrix for graph \(G_i\), \( W_i \in R^{C_i \times d} \) and \( b_i \in R^d \) are the learnable parameters of the projection layer,~$ X_i' \in R^{B \times T \times N_i \times d}$ is the projected feature representation in the unified space. After transformation, we concatenate the projected node features along the node dimension to obtain a unified representation, by:
\begin{equation}
    Z_{p} = concat(X'_{1}, X'_{2}, X'_{3}) \in \mathbb{R}^{B \times T \times N_{\text{total}} \times d}
\end{equation}
, where \(N_{\text{total}} = N_1 + N_2 + N_3\) represents the total number of nodes across all graphs. This unified representation serves as the input to subsequent graph reasoning modules, ensuring effective cross-modal information alignment. 

Then we compute an attention-weighted message passing scheme that allows nodes to exchange information across different graph structures shown in Fig.~\ref{fig:overall-framework}-(b). We first apply a Difference Operator via a temporal convolution to approximate a central difference over a window of size \( 2k+1 \):
\begin{equation}
\Delta X_t = X_t - \frac{1}{2k+1} \sum_{\tau = t-k}^{t+k} X_\tau
\end{equation}
The output \( \Delta X \) captures short-term signal deviations, fusing with original features in the Multi-Head Attention Block. For each node pair \( (i, j) \), the attention energy is computed using a difference-enhanced representation, which is formulated as:
\begin{equation}
\text{Energy}_{ij} = \phi \left( [X_i \ominus \Delta X_i \; \| \; X_j \ominus \Delta X_j] \right)
\end{equation}
\begin{equation}
\text{Attn}_{ij} = \text{Softmax} \left( \frac{\text{Energy}_{ij}}{\sqrt{d}} \right)
\end{equation}
Here, \( \ominus \) denotes element-wise subtraction, \( \| \) represents feature concatenation, and \( \phi \) is a learnable linear projection. The resulting attention matrix is combined with a prior graph structure \( A_{\text{static}} \) using a learnable fusion coefficient \( \lambda \in [0, 1] \) to obtain a dynamic adjacency matrix:
\begin{equation}
\tilde{A} = \lambda A_{\text{static}} + (1 - \lambda) \cdot \tilde{Attn}
\end{equation}
Subsequently, standard Transformer operations, including residual connection, feed-forward network, and layer normalization, are applied to restore temporal consistency and enhance expressive capacity. Finally, the output of the Difference Attention Encoder is denoted as
\begin{equation}
    Z_{p} = DAE(x_{eye}, x_{head}, x_{phy})
\end{equation}
, where $Z_{p}$ here is the fused embedding across all non-invasive modalities, which serves as the personalized representation for subsequent alignment

\subsection{Video Progressive Temporal Segment Network}
To obtain a compact representation of video content, we extract the intermediate feature vector $f_{v} \in \mathbb{R}^d$ from the penultimate layer of the PP-TSN (PaddlePaddle Temporal Segment Network) architecture. Specifically, the input video is processed by the spatial-temporal backbone network, followed by global average pooling across both temporal and spatial dimensions:
\begin{equation}
    f_{v} = \frac{1}{T}\sum_{t=1}^{T}\phi(I_{t})
\end{equation}
, where $\phi(\cdot)$ denotes the feature transformation learned by the backbone network, and 
$I_{t}$ represents the t-th temporal segment of the video.

This feature embedding $f$ is subsequently projected into the shared latent space and aligned with non-invasive data-based personalized representations via a cross-modal alignment objective, enabling personalized cybersickness prediction based solely on video input during inference.

\subsection{Cross-modal Alignment Module}
To bridge the modality gap between video and non-invasive data inputs, we introduce a cross-modal alignment objective that encourages the video embedding $z_v$ to approximate the non-invasive data-derived individual embedding $z_p$. A linear projection head maps the extracted feature $f$ into the shared embedding space:
to obtain the aligned video representation:
\begin{equation}
z_v = W_v f + b_v
\end{equation}
, where $W_v$ and $b_v$ are learnable parameters. The personality embedding $z_p$ is computed from the non-invasive signal encoder, described in Section~\ref{DAE}. The alignment is forced via mean squared error:
\begin{equation}
\mathcal{L}_{\text{align}} = \frac{1}{N} \sum_{i=1}^{N} \left| z_v^{(i)} - z_p^{(i)} \right|^2_2
\end{equation}
This alignment loss $\mathcal{L}_{\text{align}}$ serves as a supervisory signal that guides the video encoder to learn personalized representations grounded in sensor-derived embeddings. During inference, only the video encoder is retained, allowing the model to make real-time predictions without requiring any sensor input.

\subsection{Loss Definition}
The overall training objective combines a classification loss and a representation regularization term to guide both prediction accuracy and personalization:
\begin{equation}
    \mathcal{L} = \mathcal{L}_{\text{pre}} + \beta \cdot \mathcal{L}_{\text{reg}} \label{eq:total_loss}
\end{equation}
\begin{equation}
    \mathcal{L}_\text{reg} = \left\| x_c - x_{cr} \right\|_2^2 \label{eq:reg_loss}
\end{equation}
Here, $\mathcal{L}$ denotes the overall training objective that jointly optimizes classification accuracy and preserves personalized traits through feature regularization. $\mathcal{L}_{\text{pre}}$ denotes the standard cross-entropy loss between the predicted cybersickness levels and ground-truth labels. The regularization term $\mathcal{L}_{\text{reg}}$ encourages consistency between the video-based personality embedding $x_c$ and the reference representation $x_{cr}$ obtained from non-invasive signals. The hyperparameter $\beta$ controls the trade-off between personalization preservation and classification performance.

\section{Experiments}
In this section, we evaluate the personalization, performance and deployability of our proposed MM-PTGNN framework. We begin by introducing the dataset and preprocessing procedures. We then assess the representation quality of non-invasive signals and the alignment between them and video embeddings. Next, we compare prediction performance under various modality configurations and evaluate real-time feasibility. Finally, we perform ablation studies to analyze the contributions of key architectural components.

\subsection{Dataset and Preprocessing}
\begin{figure}[htbp]
    \centering
    \includegraphics[width=0.99\linewidth]{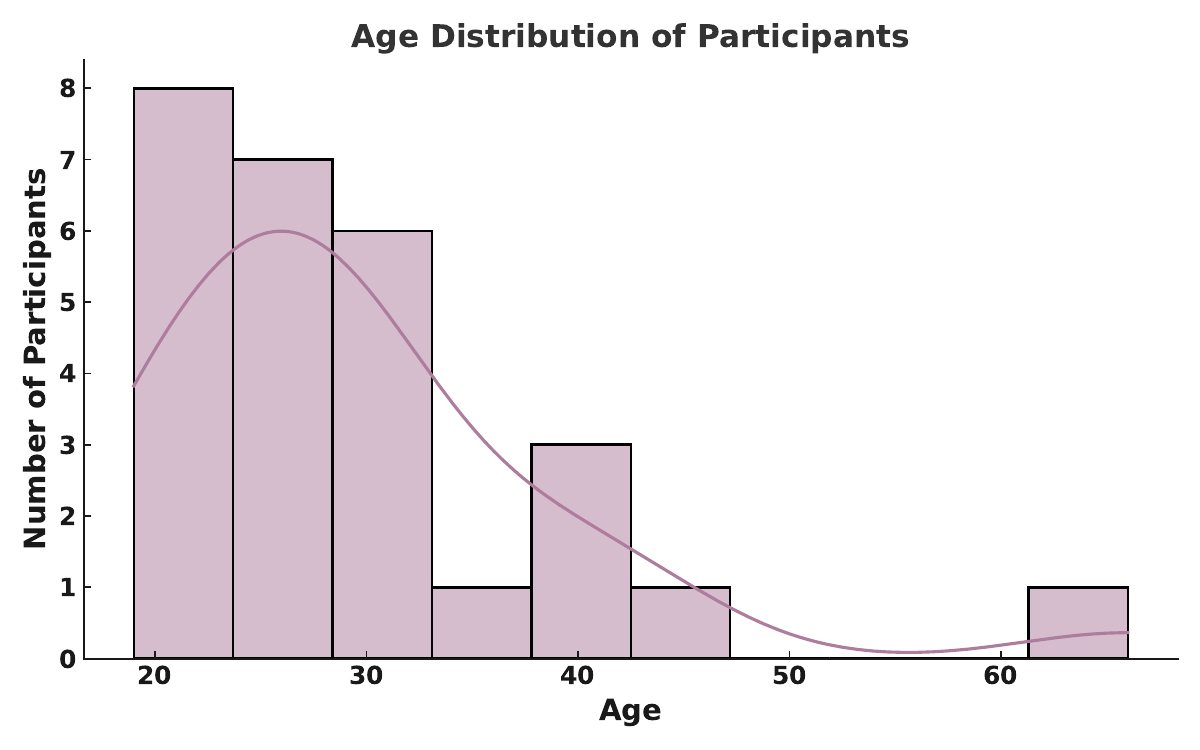}
    \caption{Overview of the dataset}
    \label{fig:dataset-overview}
\end{figure}

The dataset used in this work is sourced from one previous work~\cite{kundu2023lite}, which includes synchronized video and time-series data. The latter comprises head and eye motion (sampled at 30Hz) and physiological signals such as HR, BMP, EDA, and SKT (sampled at 1Hz) shown in Tab.~\ref{tab:the overview of the dataset}. To ensure consistency, we downsample the motion signals to 1Hz and compute statistical features over fixed intervals. Video frames are extracted at regular intervals to accelerate training. Data were collected from 27 participants (14 female, 13 male) with an average age of 29.44 years (SD: 9.95), and the detailed information is shown in Fig.~\ref{fig:dataset-overview}.

\begin{table}[h]
\centering
\caption{Overview of the dataset}
\label{tab:the overview of the dataset}
    \begin{tabular}{ccc}
    \toprule
    \textbf{Modality} & \textbf{Type (Num)} & \textbf{Sampling Rate} \\
    \midrule
    \textbf{Video}         & Video frames (1)           & 30 FPS \\
    \textbf{Head Motion}   & IMU (12)                   & 30 Hz  \\
    \textbf{Eye Motion}    & Eye tracking (38)          & 30 Hz  \\
    \textbf{Physiological} & EDA, BVP, SKT (3)          & 1 Hz   \\
    \textbf{Level}         & 0-10                       & 1      \\
    \bottomrule
    \end{tabular}
\end{table}

\subsection{Evaluation Metrics}
To evaluate the performance of cybersickness prediction under various settings, we report multiple metrics to capture overall accuracy and robustness across classes. Specifically, we use:
\textbf{Top-1 Accuracy:} the percentage of correctly predicted samples among all test samples.
\textbf{Top-3 Accuracy:} the percentage of samples for which the ground truth label appears among the top three predicted classes. This metric reflects the model's potential in ranking close categories in multi-class settings.
\textbf{Macro-F1 Score:} the unweighted average of F1 scores computed independently for each class. It is especially important in imbalanced multi-class scenarios, as it treats all classes equally regardless of their frequency.
\textbf{Cosine Similarity \& MSE (for alignment evaluation):} To assess the feature-level alignment between sensor and video embeddings, we compute the cosine similarity and mean squared error between paired representations. We report all metrics on the held-out test set averaged over \textbf{5-fold cross-validation} unless otherwise noted.

\subsection{Effectiveness of Non-invasive Embedding and Cross-modal Alignment}
\subsubsection{Non-invasive Signals Embedding Quality}
\begin{figure}
    \centering
    \includegraphics[width=0.8\linewidth]{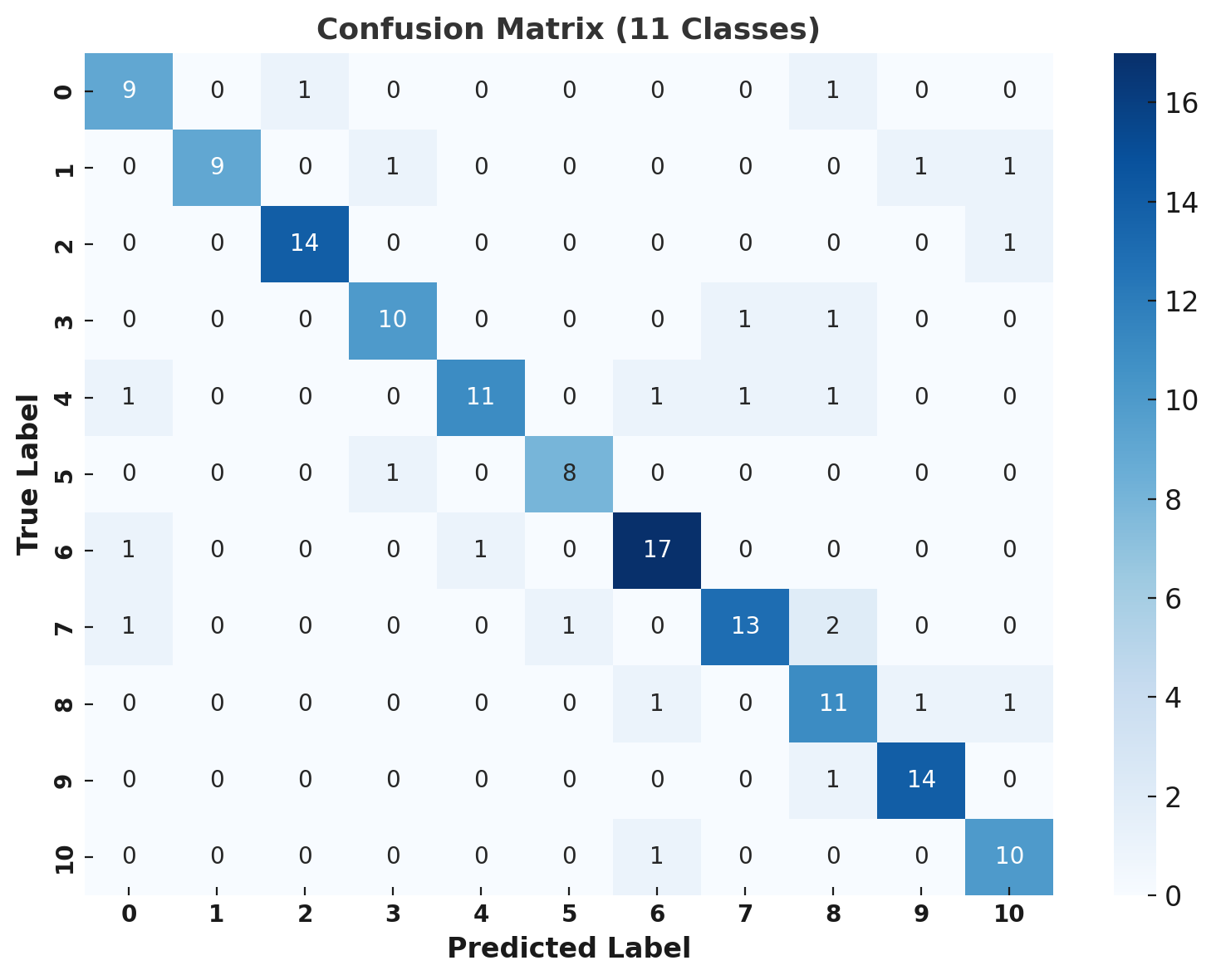}
    \caption{Confusion matrix on non-invasive signal}
    \label{fig:confusion_matrix}
\end{figure}

To verify that our model captures personalized information from non-invasive signals, we extract embeddings $z_p$ from the non-invasive signal encoder, train a lightweight classifier to predict cybersickness levels, and visualize its prediction outcomes using a confusion matrix shown in Fig.~\ref{fig:confusion_matrix}. Compared to the accuracy of 87.46\% with EEG data, our model demonstrates high classification accuracy with 84.24\%, with predictions predominantly aligned along the diagonal. This confirms that our encoder successfully captures personalized cybersickness patterns from physiological and behavioral signals.

\subsubsection{Cross-modal Feature Alignment}

To evaluate whether the video encoder effectively learns personalized representations aligned with those derived from non-invasive sensor signals, we measure the similarity between the video embedding $z_v$ and the sensor embedding $z_p$ at the feature level. Specifically, we compute the cosine similarity and mean squared error (MSE) between each embedding pair $(z_v, z_p)$ across the validation set.
To assess the importance of the cross-modal alignment objective, we set the baseline by removing the alignment loss during training. This leads to a substantial drop in similarity scores and classification accuracy, confirming that alignment is critical for transferring personalized information into the video encoder.
As shown in Tab.~\ref{tab:alignment_metrics}, the full model yields significantly higher cosine similarity and lower MSE than the variant without alignment, demonstrating the effectiveness of our joint training strategy in bridging the modality gap. As a sanity check, we also introduce a randomized baseline where sensor embeddings \( z_p \) are shuffled across samples, breaking the subject-wise alignment with \( z_v \). This results in significantly lower cosine similarity and higher MSE, confirming that the observed alignment is not due to random correlations.
\begin{table}[h]
\centering
\caption{Feature-level similarity between the embeddings}
\label{tab:alignment_metrics}
\begin{tabular}{lcc}
\toprule
\textbf{Model Variant} & \textbf{Cosine Similarity}~(\(\uparrow\)) & \textbf{MSE}~(\(\downarrow\)) \\
\midrule
Ours (Full) & \textbf{0.824} & \textbf{0.102} \\
w/o Alignment Loss & 0.611 & 0.494 \\
Randomized Baseline & 0.127 & 0.732 \\
\bottomrule
\end{tabular}
\end{table}
We also observe that replacing DiffAttention with standard graph convolution slightly degrades the alignment quality (see Section 4.3.3), suggesting that high-quality sensor embeddings are crucial for effective cross-modal representation learning.

\subsection{Cybersickness Prediction Performance}

\subsubsection{Modality-Aware Prediction Comparison}
To further investigate the effectiveness of the learned representations, we compare the predictive performance of different input modalities under the same cybersickness classification task. In particular, we evaluate three configurations: (1) using sensor embeddings \(z_p\) only, (2) using video embeddings \(z_v\) only, and (3) using video embeddings trained without alignment supervision.

The sensor-only configuration serves as an upper bound for performance using non-invasive physiological signals, while the video-only model represents the final deployable form of our framework. The version without alignment provides a baseline to assess the contribution of our cross-modal alignment mechanism.

As shown in Tab.~\ref{tab:modal_config_comparison}, the video-only model with alignment achieves performance comparable to the sensor-only model, despite requiring no sensor input during inference. In contrast, the model trained without alignment suffers from a clear performance drop. These results demonstrate that our joint training strategy successfully transfers personalized information from sensor data into the video embedding space, enabling strong predictive performance in a fully sensor-free inference setting.

\begin{table}[!htbp]
    \centering
    \caption{Performance comparison under various modality configurations.}
    \label{tab:modal_config_comparison}
    \begin{tabular}{lccc}
        \toprule
        \textbf{Modality} & \textbf{Input} & \textbf{Top-1 Acc (\%)} & \textbf{Macro-F1 (\%)} \\
        \midrule
        Kim~\cite{Kim2020VideoDrivenModels} & Video & 86.2 & - \\ 
        Martin~\cite{Martin2020VirtualrealitySickness} & Sensor & 87.1 & - \\
        Lee~\cite{Lee2019MotionSickness} & Video + Sensor & 83.2 & - \\
        Kim~\cite{Kim2019CybersicknessEEG} & EEG + Video & 89.16 & - \\
        \textbf{Ours} & \textbf{Video} & \textbf{88.4} & \textbf{84.2} \\
    \bottomrule
    \end{tabular}
\end{table}


\subsubsection{Real-time Deployment Feasibility}
To evaluate whether our method supports real-time deployment in consumer scenarios, we measure the inference latency and model size of several model variants. All measurements are performed on a single NVIDIA RTX 4080 GPU with batch size 1 to simulate realistic single-sample prediction. We focus on the final video-only inference stage, where no sensor input is required.

As shown in Tab.~\ref{tab:real_time}, our video-only aligned model achieves an average inference time of 12.4 ms per sample, which meets the latency requirement for real-time applications with the 30 FPS VR video. The model size is also relatively compact (42.3 MB), making it suitable for deployment on lightweight edge devices.

In contrast, multi-modal and EEG-based methods suffer from much higher inference delays due to increased model complexity and data pre-processing overhead. Our model can achieve as fast as Kundo~\cite{kundu2023lite}, which is a reduced DL(MLP) model but with higher accuracy.
\begin{table*}[h]
\centering
\caption{Inference feasibility comparison across models (batch size = 1).}
\label{tab:real_time}
\begin{tabular}{lcccc}
\toprule
\textbf{Model} & \textbf{Inference Input} & \textbf{Inference Time (ms)} & \textbf{Model Size (MB)} \\
\midrule
Padmanaban~\cite{Padmanaban2018StereoscopicVideos} & video & 7200 & 3020\\
Kundo~\cite{kundu2023lite} & sensor & \textbf{90} & \textbf{40.12} \\ 
Islam~\cite{islam2021cybersense} & sensor + video & 820 & 523.14\\
Jeong~\cite{Jeong2021Precyse} & EEG + Sensor & 400 & 264.48\\
\textbf{Ours} & \textbf{Video} & \textbf{93.6} & \textbf{42.3}\\ 
\bottomrule
\end{tabular}
\end{table*}
\subsection{Ablation Study}
To evaluate the individual contributions of key components in our framework, we conduct an ablation study focusing on three major aspects: the difference attention mechanism (DiffAttention), the cross-modal alignment loss, and the multi-modal training configuration. All models are trained and evaluated under identical settings using the same video-only inference setup to ensure a fair comparison.

\textbf{(1) Impact of DiffAttention:} We replace our proposed difference-aware attention module with a standard graph attention network (GAT) using raw feature similarity. As shown in Tab.~\ref{tab:ablation_study}, the absence of DiffAttention results in a notable drop in performance, especially in macro-F1, suggesting that modeling dynamic variations is essential for capturing individualized temporal patterns.

\textbf{(2) Impact of Alignment Loss:} We remove the alignment objective between video embedding \(z_v\) and sensor embedding \(z_p\) during training. Without this supervision, the model fails to effectively transfer personalized representations to the video modality, leading to degraded classification accuracy and higher variance.

These results given in Tab.~\ref{tab:ablation_study}~verify that each module contributes significantly to the model's ability to capture personalized patterns and achieve robust prediction performance under video-only inference settings.

\begin{figure}[h]
    \centering
    \includegraphics[width=0.99\linewidth]{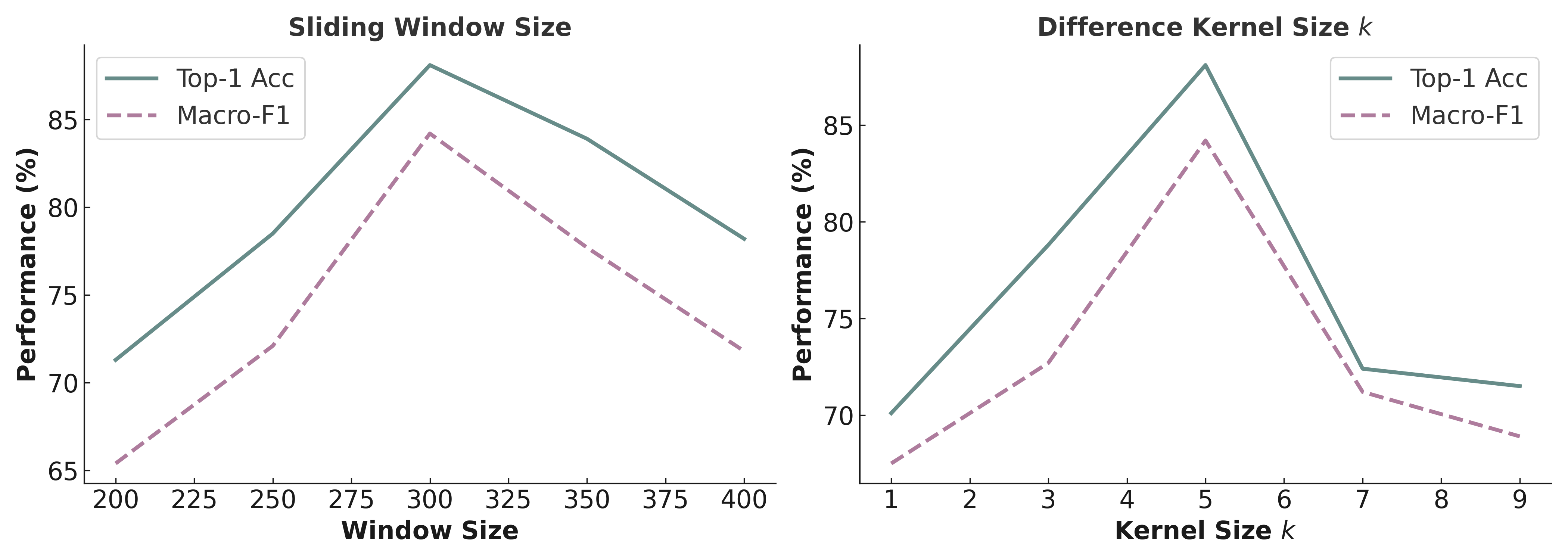}
    \caption{Impact of window size and difference kernel size}
    \label{fig:ablation_window_kernel}
\end{figure}

\textbf{(3) Impact of the training configuration:} Fig.~\ref{fig:ablation_window_kernel} illustrates the performance variation with different sliding window and kernel sizes, demonstrating that moderate temporal settings (e.g., window=300, kernel=5) yield the best prediction accuracy and stability.

\begin{table}[h]
\centering
\caption{Ablation study on the proposed framework}
\label{tab:ablation_study}
\resizebox{\columnwidth}{!}{
\begin{tabular}{lccc}
\toprule
\textbf{Model Variant} & \textbf{Top-1 Acc (\%)} & \textbf{Top-3 Acc (\%)} & \textbf{Macro-F1 (\%)} \\
\midrule
w/o DiffAttention & 70.2 & 87.3 & 64.5 \\
w/o Alignment Loss & 66.4 & 85.2 & 60.9 \\
Full Model (ours) & \textbf{73.1} & \textbf{89.9} & \textbf{68.2} \\
\bottomrule
\end{tabular}
}
\end{table}

\section{Discussion}
Due to varying individual susceptibility~\cite{li2023deep}, predicting cybersickness in consumer-grade VR applications can be challenging. Our work introduces a framework that enables real-time cybersickness prediction solely with VR video input while maintaining individual characteristics. By leveraging multi-modal non-invasive signals during training, the model learns individual variability through a Difference Attention Module and cross-modal alignment, allowing the video encoder to capture personalized traits. Experimental results demonstrate that our approach achieves performance comparable to EEG-based methods while offering greater scalability and practicality for consumer-grade VR applications.

Compared to prior works that rely heavily on EEG signals~\cite{Jeong2021Precyse, Kim2019CybersicknessEEG, Chen2020EEGCybersickness} for cybersickness prediction, our framework alleviates the dependency on intrusive biosignals, making it more suitable for real-world deployment. Integrating difference-aware attention and modality-specific graph convolution contributes significantly to the robustness and adaptability of the model across diverse user profiles. Unlike EEG-based models, which typically require complex calibration, precise electrode placement, and lengthy preprocessing pipelines, our approach uses readily accessible signals, substantially lowering the barriers to entry for widespread adoption. This methodological improvement not only ensures easier integration into consumer VR hardware but also enhances user comfort and operational feasibility.

Furthermore, our framework effectively bridges the modality gap between sensor-derived personalized embeddings and video-only representations, a crucial step toward practical and scalable deployment. By training the video encoder with signals that capture subtle physiological and behavioral differences, the model preserves critical individual-specific patterns that are typically lost in conventional single-modality video analyses. This advantage enables superior personalization and a richer understanding of how users individually respond to immersive environments, thereby improving predictive accuracy and user experience optimization.

However, the study has certain limitations. The training data, while diverse, may not encompass all possible user behaviors and environmental conditions encountered in real-world VR scenarios. For example, extreme VR content involving rapid camera movements, prolonged exposure, or interactive tasks could induce cybersickness manifestations not thoroughly represented in our current dataset. Additionally, the model's performance in entirely unseen or significantly divergent scenarios, such as novel VR applications or radically different user demographics, remains to be thoroughly evaluated.

To address these limitations, future work will focus on expanding the dataset substantially, integrating more diverse VR experiences that include a wider range of user demographics, behaviors, and task complexities. Moreover, we plan to implement adaptive alignment strategies that dynamically adjust the alignment objective according to the variability observed in real-time user interactions, thereby enhancing the model's generalization capabilities. Exploring incremental learning approaches could further refine the model's adaptability, enabling continuous updates based on user feedback in deployed scenarios.

Building upon the current framework, efforts will also be directed toward deploying the system in real-world VR applications. We will investigate efficient operation strategies specifically tailored for edge devices, such as lightweight model architectures, quantization methods, and hardware acceleration techniques, thus ensuring minimal latency and power consumption. These steps are vital for achieving a truly deployable solution capable of enhancing user comfort, immersion, and acceptance in everyday VR applications.

\section{Conclusion}
In this paper, we propose a personalized cybersickness prediction framework that enables real-time inference using only VR video. The model is first trained with multi-modal non-invasive signals, where we introduce a Difference Attention Module to enhance dynamic feature modeling. Through cross-modal alignment, personalized traits are transferred into the video encoder, making sensor-free inference possible. Experimental results show that our method achieves performance comparable to EEG-based models while offering greater scalability for consumer-grade VR applications. Our future work will focus on deploying this framework in real-world VR systems and enhancing its adaptability for practical use.

\section{Acknowledgments}
This work was supported in part by Grant No. 2025A03J3955 from the Guangzhou-HKUST(GZ) Joint Funding Scheme.

\bibliographystyle{ACM-Reference-Format}
\bibliography{sample-base}


\begin{thebibliography}{35}


\ifx \showCODEN    \undefined \def \showCODEN     #1{\unskip}     \fi
\ifx \showDOI      \undefined \def \showDOI       #1{#1}\fi
\ifx \showISBNx    \undefined \def \showISBNx     #1{\unskip}     \fi
\ifx \showISBNxiii \undefined \def \showISBNxiii  #1{\unskip}     \fi
\ifx \showISSN     \undefined \def \showISSN      #1{\unskip}     \fi
\ifx \showLCCN     \undefined \def \showLCCN      #1{\unskip}     \fi
\ifx \shownote     \undefined \def \shownote      #1{#1}          \fi
\ifx \showarticletitle \undefined \def \showarticletitle #1{#1}   \fi
\ifx \showURL      \undefined \def \showURL       {\relax}        \fi
\providecommand\bibfield[2]{#2}
\providecommand\bibinfo[2]{#2}
\providecommand\natexlab[1]{#1}
\providecommand\showeprint[2][]{arXiv:#2}

\bibitem[Balasubramanian and Soundararajan(2019)]%
        {Balasubramanian2019Egomotion}
\bibfield{author}{\bibinfo{person}{Suprith Balasubramanian} {and} \bibinfo{person}{Rajiv Soundararajan}.} \bibinfo{year}{2019}\natexlab{}.
\newblock \showarticletitle{Prediction of Discomfort due to Egomotion in Immersive Videos for Virtual Reality}. In \bibinfo{booktitle}{\emph{2019 IEEE International Symposium on Mixed and Augmented Reality (ISMAR)}}. \bibinfo{pages}{169--177}.
\newblock
\urldef\tempurl%
\url{https://doi.org/10.1109/ISMAR.2019.000-7}
\showDOI{\tempurl}


\bibitem[Chang et~al\mbox{.}(2021)]%
        {Chang2021PredictCybersicknessGaze}
\bibfield{author}{\bibinfo{person}{Eunhee Chang}, \bibinfo{person}{Hyun-Taek Kim}, {and} \bibinfo{person}{Byounghyun Yoo}.} \bibinfo{year}{2021}\natexlab{}.
\newblock \showarticletitle{Predicting cybersickness based on user's gaze behaviors in HMD-based virtual reality}.
\newblock \bibinfo{journal}{\emph{Journal of Computational Design and Engineering}}  \bibinfo{volume}{8} (\bibinfo{date}{04} \bibinfo{year}{2021}), \bibinfo{pages}{728--739}.
\newblock
\urldef\tempurl%
\url{https://doi.org/10.1093/jcde/qwab010}
\showDOI{\tempurl}


\bibitem[Chen and Lin(2020)]%
        {Chen2020EEGCybersickness}
\bibfield{author}{\bibinfo{person}{Y. Chen} {and} \bibinfo{person}{C.~T. Lin}.} \bibinfo{year}{2020}\natexlab{}.
\newblock \showarticletitle{Challenges in Deploying EEG-Based Cybersickness Detection for Consumer VR Applications}.
\newblock \bibinfo{journal}{\emph{Frontiers in Human Neuroscience}}  \bibinfo{volume}{14} (\bibinfo{year}{2020}), \bibinfo{pages}{567}.
\newblock


\bibitem[Choi et~al\mbox{.}(2024)]%
        {choi2024early}
\bibfield{author}{\bibinfo{person}{Yoonseon Choi}, \bibinfo{person}{Dayoung Jeong}, \bibinfo{person}{Bogoan Kim}, {and} \bibinfo{person}{Kyungsik Han}.} \bibinfo{year}{2024}\natexlab{}.
\newblock \showarticletitle{Early Prediction of Cybersickness in Virtual Reality Using a Large Language Model for Multimodal Time Series Data}. In \bibinfo{booktitle}{\emph{Companion of the 2024 on ACM International Joint Conference on Pervasive and Ubiquitous Computing}}. \bibinfo{pages}{25--29}.
\newblock


\bibitem[Fan et~al\mbox{.}(2025)]%
        {fan2025medgnnmultiresolutionspatiotemporalgraph}
\bibfield{author}{\bibinfo{person}{Wei Fan}, \bibinfo{person}{Jingru Fei}, \bibinfo{person}{Dingyu Guo}, \bibinfo{person}{Kun Yi}, \bibinfo{person}{Xiaozhuang Song}, \bibinfo{person}{Haolong Xiang}, \bibinfo{person}{Hangting Ye}, {and} \bibinfo{person}{Min Li}.} \bibinfo{year}{2025}\natexlab{}.
\newblock \bibinfo{title}{MedGNN: Towards Multi-resolution Spatiotemporal Graph Learning for Medical Time Series Classification}.
\newblock
\newblock
\showeprint[arxiv]{2502.04515}~[cs.LG]
\urldef\tempurl%
\url{https://arxiv.org/abs/2502.04515}
\showURL{%
\tempurl}


\bibitem[Fernandes and Feiner(2016)]%
        {fernandes2016combating}
\bibfield{author}{\bibinfo{person}{A.S. Fernandes} {and} \bibinfo{person}{S.K. Feiner}.} \bibinfo{year}{2016}\natexlab{}.
\newblock \showarticletitle{Combating VR sickness through subtle dynamic field-of-view modification}. In \bibinfo{booktitle}{\emph{Proc. IEEE Symp. 3D User Interfaces (3DUI)}}. \bibinfo{pages}{201--210}.
\newblock
\urldef\tempurl%
\url{https://doi.org/10.1109/3DUI.2016.7460053}
\showDOI{\tempurl}


\bibitem[Gavgani et~al\mbox{.}(2018)]%
        {gavgani2018cybersickness}
\bibfield{author}{\bibinfo{person}{A.M. Gavgani}, \bibinfo{person}{K.V. Nesbitt}, \bibinfo{person}{K.L. Blackmore}, {and} \bibinfo{person}{E. Nalivaiko}.} \bibinfo{year}{2018}\natexlab{}.
\newblock \showarticletitle{Cybersickness-related changes in brain hemodynamics: A pilot study comparing VR and non-VR tasks}. In \bibinfo{booktitle}{\emph{Proc. IEEE Int. Symp. Mixed Augmented Reality (ISMAR)}}. \bibinfo{pages}{121--126}.
\newblock
\urldef\tempurl%
\url{https://doi.org/10.1109/ISMAR.2018.00036}
\showDOI{\tempurl}


\bibitem[Islam et~al\mbox{.}(2021a)]%
        {islam2021cybersense}
\bibfield{author}{\bibinfo{person}{Rifatul Islam}, \bibinfo{person}{Samuel Ang}, {and} \bibinfo{person}{John Quarles}.} \bibinfo{year}{2021}\natexlab{a}.
\newblock \showarticletitle{Cybersense: A closed-loop framework to detect cybersickness severity and adaptively apply reduction techniques}. In \bibinfo{booktitle}{\emph{2021 IEEE Conference on virtual reality and 3d user interfaces abstracts and workshops (VRW)}}. IEEE, \bibinfo{pages}{148--155}.
\newblock
\urldef\tempurl%
\url{https://doi.org/10.1109/VRW52623.2021.00035}
\showDOI{\tempurl}


\bibitem[Islam et~al\mbox{.}(2021b)]%
        {Islam2021IntegratedHMDsSensors}
\bibfield{author}{\bibinfo{person}{Rifatul Islam}, \bibinfo{person}{Kevin Desai}, {and} \bibinfo{person}{John Quarles}.} \bibinfo{year}{2021}\natexlab{b}.
\newblock \showarticletitle{Cybersickness Prediction from Integrated HMD’s Sensors: A Multimodal Deep Fusion Approach using Eye-tracking and Head-tracking Data}. In \bibinfo{booktitle}{\emph{2021 IEEE International Symposium on Mixed and Augmented Reality (ISMAR)}}. \bibinfo{pages}{31--40}.
\newblock
\urldef\tempurl%
\url{https://doi.org/10.1109/ISMAR52148.2021.00017}
\showDOI{\tempurl}


\bibitem[Jeon et~al\mbox{.}(2023)]%
        {JEON2023104929}
\bibfield{author}{\bibinfo{person}{Jin~Yong Jeon}, \bibinfo{person}{Hyun~In Jo}, {and} \bibinfo{person}{Kounseok Lee}.} \bibinfo{year}{2023}\natexlab{}.
\newblock \showarticletitle{Psycho-physiological restoration with audio-visual interactions through virtual reality simulations of soundscape and landscape experiences in urban, waterfront, and green environments}.
\newblock \bibinfo{journal}{\emph{Sustainable Cities and Society}}  \bibinfo{volume}{99} (\bibinfo{year}{2023}), \bibinfo{pages}{104929}.
\newblock
\showISSN{2210-6707}
\urldef\tempurl%
\url{https://doi.org/10.1016/j.scs.2023.104929}
\showDOI{\tempurl}


\bibitem[Jeong and Han(2022)]%
        {jeong2022leveraging}
\bibfield{author}{\bibinfo{person}{Dayoung Jeong} {and} \bibinfo{person}{Kyungsik Han}.} \bibinfo{year}{2022}\natexlab{}.
\newblock \showarticletitle{Leveraging multimodal sensory information in cybersickness prediction}. In \bibinfo{booktitle}{\emph{Proceedings of the 28th ACM Symposium on Virtual Reality Software and Technology}}. \bibinfo{pages}{1--2}.
\newblock


\bibitem[Jeong and Han(2024)]%
        {Jeong2021Precyse}
\bibfield{author}{\bibinfo{person}{Dayoung Jeong} {and} \bibinfo{person}{Kyungsik Han}.} \bibinfo{year}{2024}\natexlab{}.
\newblock \showarticletitle{PRECYSE: Predicting Cybersickness using Transformer for Multimodal Time-Series Sensor Data}.
\newblock \bibinfo{journal}{\emph{Proc. ACM Interact. Mob. Wearable Ubiquitous Technol.}} \bibinfo{volume}{8}, \bibinfo{number}{2}, Article \bibinfo{articleno}{42} (\bibinfo{date}{May} \bibinfo{year}{2024}), \bibinfo{numpages}{24}~pages.
\newblock
\urldef\tempurl%
\url{https://doi.org/10.1145/3659594}
\showDOI{\tempurl}


\bibitem[Jeong et~al\mbox{.}(2019)]%
        {Jeong2019DeepLearningAlgorithms}
\bibfield{author}{\bibinfo{person}{Daekyo Jeong}, \bibinfo{person}{Sangbong Yoo}, {and} \bibinfo{person}{Jang Yun}.} \bibinfo{year}{2019}\natexlab{}.
\newblock \showarticletitle{Cybersickness Analysis with EEG Using Deep Learning Algorithms}. In \bibinfo{booktitle}{\emph{2019 IEEE Conference on Virtual Reality and 3D User Interfaces (VR)}}. \bibinfo{pages}{827--835}.
\newblock
\urldef\tempurl%
\url{https://doi.org/10.1109/VR.2019.8798334}
\showDOI{\tempurl}


\bibitem[Jiang et~al\mbox{.}(2024)]%
        {Jiang2024ChemistryVR}
\bibfield{author}{\bibinfo{person}{Guanxuan Jiang}, \bibinfo{person}{Xuansheng Xia}, \bibinfo{person}{Yue Li}, \bibinfo{person}{Hai-Ning Liang}, {and} \bibinfo{person}{Pan Hui}.} \bibinfo{year}{2024}\natexlab{}.
\newblock \showarticletitle{ChemistryVR: Enhancing Educational Experiences through Virtual Chemistry Lab Simulations}. In \bibinfo{booktitle}{\emph{SIGGRAPH Asia 2024 Educator's Forum}} \emph{(\bibinfo{series}{SA '24})}. \bibinfo{publisher}{Association for Computing Machinery}, \bibinfo{address}{New York, NY, USA}, Article \bibinfo{articleno}{1}, \bibinfo{numpages}{5}~pages.
\newblock
\showISBNx{9798400711367}
\urldef\tempurl%
\url{https://doi.org/10.1145/3680533.3697068}
\showDOI{\tempurl}


\bibitem[Jin et~al\mbox{.}(2023)]%
        {jin2023development}
\bibfield{author}{\bibinfo{person}{Shan Jin}, \bibinfo{person}{Yuyang Wang}, \bibinfo{person}{Lik-Hang Lee}, \bibinfo{person}{Xinyi Luo}, {and} \bibinfo{person}{Pan Hui}.} \bibinfo{year}{2023}\natexlab{}.
\newblock \showarticletitle{Development of an immersive simulator for improving student chemistry learning efficiency}. In \bibinfo{booktitle}{\emph{Proceedings of the 16th International Symposium on Visual Information Communication and Interaction}}. \bibinfo{pages}{1--8}.
\newblock
\urldef\tempurl%
\url{https://doi.org/10.1145/3615522.36155}
\showDOI{\tempurl}


\bibitem[Kim and Ro(2020)]%
        {Kim2020VideoDrivenModels}
\bibfield{author}{\bibinfo{person}{H.~G. Kim} {and} \bibinfo{person}{Y.~M. Ro}.} \bibinfo{year}{2020}\natexlab{}.
\newblock \showarticletitle{Limitations of Optical Flow Features for VR Sickness Prediction in Dynamic Environments}. In \bibinfo{booktitle}{\emph{IEEE International Conference on Multimedia and Expo (ICME)}}. \bibinfo{pages}{1–6}.
\newblock


\bibitem[Kim et~al\mbox{.}(2019a)]%
        {Kim2019CybersicknessEEG}
\bibfield{author}{\bibinfo{person}{J. Kim}, \bibinfo{person}{W. Kim}, \bibinfo{person}{H. Oh}, \bibinfo{person}{S. Lee}, {and} \bibinfo{person}{S. Lee}.} \bibinfo{year}{2019}\natexlab{a}.
\newblock \showarticletitle{A Deep Cybersickness Predictor Based on Brain Signal Analysis for Virtual Reality Contents}. In \bibinfo{booktitle}{\emph{IEEE/CVF International Conference on Computer Vision (ICCV)}}. \bibinfo{pages}{10579–10588}.
\newblock


\bibitem[Kim et~al\mbox{.}(2019b)]%
        {Kim2019BrainSignalAnalysis}
\bibfield{author}{\bibinfo{person}{Jinwoo Kim}, \bibinfo{person}{Woojae Kim}, \bibinfo{person}{Heeseok Oh}, \bibinfo{person}{Seongmin Lee}, {and} \bibinfo{person}{Sanghoon Lee}.} \bibinfo{year}{2019}\natexlab{b}.
\newblock \showarticletitle{A Deep Cybersickness Predictor Based on Brain Signal Analysis for Virtual Reality Contents}. In \bibinfo{booktitle}{\emph{2019 IEEE/CVF International Conference on Computer Vision (ICCV)}}. \bibinfo{pages}{10579--10588}.
\newblock
\urldef\tempurl%
\url{https://doi.org/10.1109/ICCV.2019.01068}
\showDOI{\tempurl}


\bibitem[Kim et~al\mbox{.}(2020)]%
        {Kim2020EstimateVRSickness}
\bibfield{author}{\bibinfo{person}{Seongyeop Kim}, \bibinfo{person}{Sangmin Lee}, {and} \bibinfo{person}{Yong~Man Ro}.} \bibinfo{year}{2020}\natexlab{}.
\newblock \showarticletitle{Estimating VR Sickness Caused By Camera Shake in VR Videography}. In \bibinfo{booktitle}{\emph{2020 IEEE International Conference on Image Processing (ICIP)}}. \bibinfo{pages}{3433--3437}.
\newblock
\urldef\tempurl%
\url{https://doi.org/10.1109/ICIP40778.2020.9190721}
\showDOI{\tempurl}


\bibitem[Kundu et~al\mbox{.}(2023)]%
        {kundu2023lite}
\bibfield{author}{\bibinfo{person}{Ripan~Kumar Kundu}, \bibinfo{person}{Rifatul Islam}, \bibinfo{person}{John Quarles}, {and} \bibinfo{person}{Khaza~Anuarul Hoque}.} \bibinfo{year}{2023}\natexlab{}.
\newblock \bibinfo{title}{LiteVR: Interpretable and Lightweight Cybersickness Detection using Explainable AI}.
\newblock
\newblock
\showeprint[arxiv]{2302.03037}~[cs.HC]
\urldef\tempurl%
\url{https://arxiv.org/abs/2302.03037}
\showURL{%
\tempurl}


\bibitem[Laghari et~al\mbox{.}(2021)]%
        {laghari2021veteran}
\bibfield{author}{\bibinfo{person}{Asif Laghari}, \bibinfo{person}{Imran Niazi}, {and} \bibinfo{person}{Joseph Coughlan}.} \bibinfo{year}{2021}\natexlab{}.
\newblock \showarticletitle{VR Veterans vs. Novices: How Prior Experience Shapes Cybersickness and Adaptation}. In \bibinfo{booktitle}{\emph{Proc. IEEE Conf. Virtual Real. 3D User Interfaces (IEEE VR)}}. \bibinfo{pages}{123--132}.
\newblock
\urldef\tempurl%
\url{https://doi.org/10.1109/VRW52623.2021.00030}
\showDOI{\tempurl}


\bibitem[Lee et~al\mbox{.}(2019)]%
        {Lee2019MotionSickness}
\bibfield{author}{\bibinfo{person}{Tae~Min Lee}, \bibinfo{person}{Jong-Chul Yoon}, {and} \bibinfo{person}{In-Kwon Lee}.} \bibinfo{year}{2019}\natexlab{}.
\newblock \showarticletitle{Motion Sickness Prediction in Stereoscopic Videos using 3D Convolutional Neural Networks}.
\newblock \bibinfo{journal}{\emph{IEEE Transactions on Visualization and Computer Graphics}} \bibinfo{volume}{25}, \bibinfo{number}{5} (\bibinfo{year}{2019}), \bibinfo{pages}{1919--1927}.
\newblock
\urldef\tempurl%
\url{https://doi.org/10.1109/TVCG.2019.2899186}
\showDOI{\tempurl}


\bibitem[Li et~al\mbox{.}(2023)]%
        {li2023deep}
\bibfield{author}{\bibinfo{person}{Ruichen Li}, \bibinfo{person}{Yuyang Wang}, \bibinfo{person}{Handi Yin}, \bibinfo{person}{Jean-R{\'e}my Chardonnet}, {and} \bibinfo{person}{Pan Hui}.} \bibinfo{year}{2023}\natexlab{}.
\newblock \showarticletitle{A deep cybersickness predictor through kinematic data with encoded physiological representation}. In \bibinfo{booktitle}{\emph{2023 IEEE International Symposium on Mixed and Augmented Reality (ISMAR)}}. IEEE, \bibinfo{pages}{1132--1141}.
\newblock


\bibitem[Martin et~al\mbox{.}(2020)]%
        {Martin2020VirtualrealitySickness}
\bibfield{author}{\bibinfo{person}{Nicolas Martin}, \bibinfo{person}{Nicolas Mathieu}, \bibinfo{person}{pallamin nico}, \bibinfo{person}{Martin Ragot}, {and} \bibinfo{person}{Diverrez J-Marc}.} \bibinfo{year}{2020}\natexlab{}.
\newblock \showarticletitle{Virtual reality sickness detection: an approach based on physiological signals and machine learning}.
\newblock
\urldef\tempurl%
\url{https://doi.org/10.1109/ISMAR50242.2020.00065}
\showDOI{\tempurl}


\bibitem[Padmanaban et~al\mbox{.}(2018)]%
        {Padmanaban2018StereoscopicVideos}
\bibfield{author}{\bibinfo{person}{Nitish Padmanaban}, \bibinfo{person}{Timon Ruban}, \bibinfo{person}{Vincent Sitzmann}, \bibinfo{person}{Anthony~M. Norcia}, {and} \bibinfo{person}{Gordon Wetzstein}.} \bibinfo{year}{2018}\natexlab{}.
\newblock \showarticletitle{Towards a Machine-Learning Approach for Sickness Prediction in 360° Stereoscopic Videos}.
\newblock \bibinfo{journal}{\emph{IEEE Transactions on Visualization and Computer Graphics}} \bibinfo{volume}{24}, \bibinfo{number}{4} (\bibinfo{year}{2018}), \bibinfo{pages}{1594--1603}.
\newblock
\urldef\tempurl%
\url{https://doi.org/10.1109/TVCG.2018.2793560}
\showDOI{\tempurl}


\bibitem[Padmanaban et~al\mbox{.}(2019)]%
        {Padmanaban2019FOVReduction}
\bibfield{author}{\bibinfo{person}{N. Padmanaban}, \bibinfo{person}{T. Ruban}, {and} \bibinfo{person}{G. Wetzstein}.} \bibinfo{year}{2019}\natexlab{}.
\newblock \showarticletitle{Dynamic Field-of-View Restriction for Cybersickness Reduction in VR}. In \bibinfo{booktitle}{\emph{Proceedings of the ACM SIGGRAPH Symposium on Applied Perception}}. \bibinfo{pages}{1–8}.
\newblock


\bibitem[Qiu et~al\mbox{.}(2023)]%
        {qiu2023DifferAttention}
\bibfield{author}{\bibinfo{person}{Xuanjie Qiu}, \bibinfo{person}{Fang Yan}, {and} \bibinfo{person}{Haihong Liu}.} \bibinfo{year}{2023}\natexlab{}.
\newblock \showarticletitle{A difference attention ResNet-LSTM network for epileptic seizure detection using EEG signal}.
\newblock \bibinfo{journal}{\emph{Biomedical Signal Processing and Control}}  \bibinfo{volume}{83} (\bibinfo{year}{2023}), \bibinfo{pages}{104652}.
\newblock
\showISSN{1746-8094}
\urldef\tempurl%
\url{https://doi.org/10.1016/j.bspc.2023.104652}
\showDOI{\tempurl}


\bibitem[Ramaseri-Chandra and Reza(2025)]%
        {ramaseri2025dynamic}
\bibfield{author}{\bibinfo{person}{Ananth~N. Ramaseri-Chandra} {and} \bibinfo{person}{Hassan Reza}.} \bibinfo{year}{2025}\natexlab{}.
\newblock \bibinfo{title}{Dynamic Cybersickness Mitigation via Adaptive FFR and FoV adjustments}.
\newblock
\newblock
\showeprint[arxiv]{2502.03419}~[cs.HC]
\urldef\tempurl%
\url{https://arxiv.org/abs/2502.03419}
\showURL{%
\tempurl}


\bibitem[Rebenitsch and Owen(2016)]%
        {rebenitsch2016individual}
\bibfield{author}{\bibinfo{person}{Lisa Rebenitsch} {and} \bibinfo{person}{Charles Owen}.} \bibinfo{year}{2016}\natexlab{}.
\newblock \showarticletitle{Individual Variation in Susceptibility to Cybersickness}.
\newblock \bibinfo{journal}{\emph{ACM Transactions on Applied Perception}} \bibinfo{volume}{13}, \bibinfo{number}{3} (\bibinfo{year}{2016}), \bibinfo{pages}{1--23}.
\newblock
\urldef\tempurl%
\url{https://doi.org/10.1145/2912125}
\showDOI{\tempurl}


\bibitem[Riccio and Stoffregen(1991)]%
        {riccio1991ecological}
\bibfield{author}{\bibinfo{person}{Gary~E Riccio} {and} \bibinfo{person}{Thomas~A Stoffregen}.} \bibinfo{year}{1991}\natexlab{}.
\newblock \showarticletitle{An ecological theory of motion sickness and postural instability}.
\newblock \bibinfo{journal}{\emph{Ecological Psychology}} \bibinfo{volume}{3}, \bibinfo{number}{3} (\bibinfo{year}{1991}), \bibinfo{pages}{195--240}.
\newblock
\urldef\tempurl%
\url{https://doi.org/10.1207/s15326969eco0303_2}
\showDOI{\tempurl}


\bibitem[Smith and Lee(2021)]%
        {Smith2021VisualVestibularConflict}
\bibfield{author}{\bibinfo{person}{J. Smith} {and} \bibinfo{person}{S. Lee}.} \bibinfo{year}{2021}\natexlab{}.
\newblock \showarticletitle{Mitigating Visually Induced Motion Sickness Through Adaptive Motion Compensation in Virtual Environments}.
\newblock \bibinfo{journal}{\emph{IEEE Transactions on Visualization and Computer Graphics}} \bibinfo{volume}{28}, \bibinfo{number}{4} (\bibinfo{year}{2021}), \bibinfo{pages}{2123–2132}.
\newblock


\bibitem[Stanney et~al\mbox{.}(2020)]%
        {stanney2020sex}
\bibfield{author}{\bibinfo{person}{Kay~M. Stanney}, \bibinfo{person}{Kelly~S. Hale}, \bibinfo{person}{Isabelina Nahmens}, {and} \bibinfo{person}{Robert~S. Kennedy}.} \bibinfo{year}{2020}\natexlab{}.
\newblock \showarticletitle{Gender Differences in Cybersickness: Clarifying the Role of Navigation and Interaction Modalities}.
\newblock \bibinfo{journal}{\emph{Frontiers in Virtual Reality}}  \bibinfo{volume}{1} (\bibinfo{year}{2020}), \bibinfo{pages}{572924}.
\newblock
\urldef\tempurl%
\url{https://doi.org/10.3389/frvir.2020.572924}
\showDOI{\tempurl}


\bibitem[Toyoda et~al\mbox{.}(2022)]%
        {VR-based-industries}
\bibfield{author}{\bibinfo{person}{Ryo Toyoda}, \bibinfo{person}{Fernando Russo~Abegão}, {and} \bibinfo{person}{Jarka Glassey}.} \bibinfo{year}{2022}\natexlab{}.
\newblock \showarticletitle{VR-based health and safety training in various high-risk engineering industries: a literature review}.
\newblock \bibinfo{journal}{\emph{International Journal of Educational Technology in Higher Education}}  \bibinfo{volume}{19} (\bibinfo{date}{08} \bibinfo{year}{2022}).
\newblock
\urldef\tempurl%
\url{https://doi.org/10.1186/s41239-022-00349-3}
\showDOI{\tempurl}


\bibitem[Wu et~al\mbox{.}(2024b)]%
        {Wu2024HypergraphMultiModalLLM}
\bibfield{author}{\bibinfo{person}{Minghui Wu}, \bibinfo{person}{Chenxu Zhao}, \bibinfo{person}{Anyang Su}, \bibinfo{person}{Donglin Di}, \bibinfo{person}{Tianyu Fu}, \bibinfo{person}{Da An}, \bibinfo{person}{Min He}, \bibinfo{person}{Ya Gao}, \bibinfo{person}{Meng Ma}, \bibinfo{person}{Kun Yan}, {and} \bibinfo{person}{Ping Wang}.} \bibinfo{year}{2024}\natexlab{b}.
\newblock \showarticletitle{Hypergraph Multi-modal Large Language Model: Exploiting EEG and Eye-tracking Modalities to Evaluate Heterogeneous Responses for Video Understanding}. In \bibinfo{booktitle}{\emph{Proceedings of the 32nd ACM International Conference on Multimedia}} (Melbourne VIC, Australia) \emph{(\bibinfo{series}{MM '24})}. \bibinfo{publisher}{Association for Computing Machinery}, \bibinfo{address}{New York, NY, USA}, \bibinfo{pages}{7316–7325}.
\newblock
\showISBNx{9798400706868}
\urldef\tempurl%
\url{https://doi.org/10.1145/3664647.3680810}
\showDOI{\tempurl}


\bibitem[Wu et~al\mbox{.}(2024a)]%
        {wu2024aienhancedvirtualrealitymedicine}
\bibfield{author}{\bibinfo{person}{Yixuan Wu}, \bibinfo{person}{Kaiyuan Hu}, \bibinfo{person}{Danny~Z. Chen}, {and} \bibinfo{person}{Jian Wu}.} \bibinfo{year}{2024}\natexlab{a}.
\newblock \bibinfo{title}{AI-Enhanced Virtual Reality in Medicine: A Comprehensive Survey}.
\newblock
\newblock
\showeprint[arxiv]{2402.03093}~[cs.CV]
\urldef\tempurl%
\url{https://arxiv.org/abs/2402.03093}
\showURL{%
\tempurl}


\end{thebibliography}

\appendix









\end{document}